# Exploiting Functional Dependencies in Qualitative Probabilistic Reasoning

Michael P. Wellman
AI Technology Office, Wright R&D Center
Wright-Patterson AFB, OH 45433
*wellman@wrdc.af.mil*

**Abstract**

Functional dependencies restrict the potential interactions among variables connected in a probabilistic network. This restriction can be exploited in qualitative probabilistic reasoning by introducing deterministic variables and modifying the inference rules to produce stronger conclusions in the presence of functional relations. I describe how to accomplish these modifications in qualitative probabilistic networks by exhibiting the update procedures for graphical transformations involving probabilistic and deterministic variables and combinations. A simple example demonstrates that the augmented scheme can reduce qualitative ambiguity that would arise without the special treatment of functional dependency. Analysis of qualitative synergy reveals that new higher-order relations are required to reason effectively about synergistic interactions among deterministic variables.

## 1 Introduction

A degenerate special case of probabilistic relations arises when the variables are connected by a deterministic, or *functional*, relationship. Although strictly a subclass of the general probabilistic case, it often pays to distinguish relations of this type and earmark them for special treatment. For example, Neufeld and Poole [1988] rely on the distinction between deterministic (implication) and probabilistic (confirmation) relations in their application of qualitative probability to default reasoning. The important difference between functional and probabilistic dependencies is in the restrictions they impose on potential interactions among connected variables. In this paper, I investigate the opportunity to exploit these constraints where deterministic and probabilistic variables coexist in networks of qualitative relations. The product is a special set of inference rules for manipulating combinations of these relations. Incorporating these rules in a hybrid representation scheme results in a language more expressive and powerful than would be obtained from the simple union of its deterministic and probabilistic components.

The advantage of a functional relation is that the arguments of the function completely determine (hence the term *deterministic*) its value. Any other variables added to the argument list would be superfluous. In contrast, introducing additional conditioning variables to a conditional probability can cause its value to change arbitrarily. Because deterministic relations impose stricter limits on potential interactions, they are inherently more *modular* than probabilistic relations [Heckerman and Horvitz, 1988].

A formal expression of this enhanced modularity can be found in graphical criteria for conditional independence in probabilistic networks [Pearl *et al.*, 1989]. A network containing deterministic variables entails more conditional independencies than an identical structure representing purely probabilistic relations. As demonstrated below, a similar improvement can be achieved for other qualitative properties of relations in probabilistic networks—in particular, monotonicity. Indeed, the ability to derive stronger qualitative conclusions from networks containing deterministic relations is ultimately due to the extra independencies sanctioned by the functional constraints.

Specification of deterministic variables in probabilistic network representations was introduced by Shachter for numeric influence diagrams [1988]. The hybrid representation scheme presented here extends the *qualitative probabilistic network* (QPN) formalism [Wellman, 1990a] to accommodate deterministic relations. Its manipulation of functional dependencies draws on the work of Michelena and Agogino [1989] on deterministic monotonic influence diagrams (dMIDs). In addition, the synthesis yields new inference rules not expressible in QPNs or dMIDs alone.

## 2 Example: Reducing Ambiguity

A probabilistic network (also called a belief network or influence diagram, with some variations) is a directed acyclic graph composed of nodes denoting random variables and edges indicating their probabilistic dependencies. A network represents a valid dependency structure, called an *Independence-* or *I-map* [Pearl et al., 1989], if the joint distribution over the entire variable set can be factored into the conditional probabilities of each node given its predecessors. When a network is an I-map, the graphical *d-separation* criterion is a sufficient condition for conditional independence. In the presence of deterministic variables, the stronger *D-separation* (note uppercase) condition may be applied. For definitions of these conditions and thorough discussion of their properties, see the work of Pearl et al. ([1989], for example).

The probabilistic network of Figure 1 provides an example of the distinction between d- and D-separation. Nodes $x$ and $y$ may be marginally dependent, as they have a common predecessor, $w$. (The dashed inner ellipse indicates that $w$ may or may not be deterministic.) They are conditionally independent given $w$, as $w$ blocks the only path (undirected) between them. Suppose $w$ is a probabilistic node. Then $x$ and $y$ are *not* conditionally independent given $z$, because $z$ provides only probabilistic information about $w$. Even given $z$, information about $x$ is potential evidence impinging on $w$, and hence affects belief about $y$. These independence and dependence assertions are in accordance with the d-separation criterion applied to the graph.

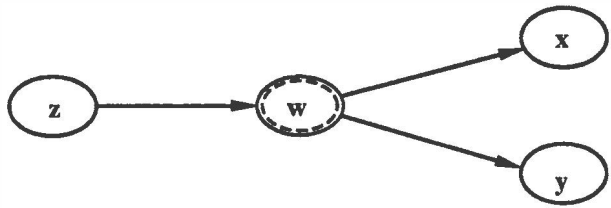

Figure 1: Nodes $x$ and $y$ are not d-separated by $\{z\}$, but are D-separated if $w$ is deterministic.

If $w$ is functionally determined by $z$, however, then knowledge of $z$ leaves no room for further influence from $x$ on belief about $w$. Therefore there is no effective path from $x$ to $y$, and the two variables are conditionally independent given $z$. This is verified by the stronger separation criterion: if $w$ is deterministic, $x$ and $y$ are D-separated by $\{z\}$.

The determinism of $w$ impacts relations beyond conditional independence. Consider the qualitative probabilistic network of Figure 2. This network is similar to the one above, with one extra edge and signs $\delta_i \in \{+, -, ?\}$ on each link indicating the *qualitative influences* holding among the variables. Qualitative influences are a type of monotonicity constraint on the probabilistic dependence between the associated variables. Inference in QPNs consists of combinations and manipulations of these influences to derive new influences among variables not directly connected in the original network.

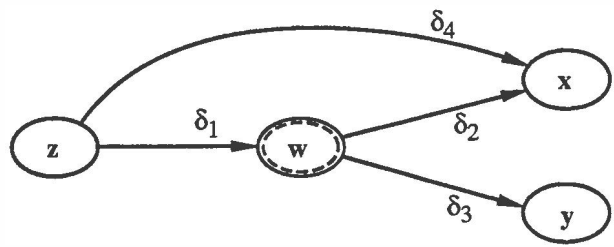

Figure 2: A qualitative probabilistic network. Its implications depend on whether $w$ is deterministic.

Suppose we are interested in determining the qualitative relation of $z$ on $y$ given $x$ in this model. In the given network, the value of $y$ is specified in terms of $w$; $z$ and $x$ are only indirectly related to $y$. To compute the relation of interest, we transform the network via a series of node reductions and link reversals until the relation is displayed directly. In our example, $z$ must replace $w$ as a predecessor of $y$.

The probabilistic semantics of qualitative influences sanctions simple graphical transformation operations based on sign multiplication ($\otimes$) and addition ($\oplus$) [Wellman, 1990a; Wellman, 1990b]. Treating the network as a standard QPN, the best transformation consists of a reversal of the link from $w$ to $x$, followed by a reduction of $w$ from the network. The resulting network is depicted in Figure 3a. Note that regardless of the signs of the original relations (as long as they are nonzero), the qualitative relation of $z$ on $y$ given $x$ is ambiguous (that is, $\delta = ?$). The ambiguity in this case is not spurious; the potential interaction of $z$ and $w$ in their relation to $x$ admits an arbitrary probabilistic relation of $z$ on $w$ given $x$. This ambiguity then propagates directly to the conditional influence of $z$ on $y$.

In contrast, if $w$ is a function of $z$, then knowing $x$ can provide no additional information, and the potential interaction of $z$ and $w$ on $x$ is irrelevant. In this case, shown in Figure 3b, the relation of $z$ on $y$ (whether $x$ is given or not) depends only on $z$'s influence on $w$ and $w$'s on $y$. Note also that this network reflects the conditional independence of $x$ and $y$ given $z$, whereas the network of Figure 3a does not.

Thus, by recognizing the special case of deterministic relations, we are able to obtain strictly stronger qualitative conclusions (unless of course $\delta_1$ or $\delta_3$ are ?, in which case the results are equivalent). Not only can we detect more independencies, we can also resolve ambiguities that are prevented by these additional independence conditions. Moreover, the example above demonstrates that this phenomenon can be manifested even when all the variables of interest are probabilistic.

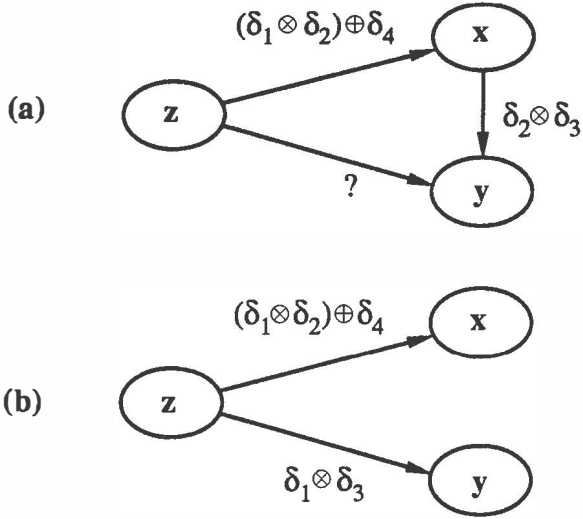

Figure 3: (a) The transformed network using standard QPN operations. (b) The result obtained by exploiting the deterministic nature of $w$.

We can achieve reductions of ambiguity without resorting to explicit identification of independencies via D-separation. In the remaining sections, I demonstrate how the advantages of functional relations can be realized locally by simple modifications to the QPN transformation operations for cases where one or more of the nodes involved are deterministic.

## 3 Qualitative Relations

### 3.1 Probabilistic Relations

QPNs support two types of qualitative probabilistic relations. Influences describe the direction of a probabilistic relation, and *synergies* describe the interaction among influences. The bulk of this analysis concerns qualitative influences; further discussion of synergy is deferred to Section 5.

A qualitative influence is a kind of probabilistic monotonicity constraint on random variables. We say that that $a$ positively influences $b$ iff the probability distribution for $b$ given $a$ is increasing in $a$, in the sense of first-order stochastic dominance (FSD). When $b$ has other predecessors besides $a$, the relation must hold for any assignment of values to those variables. In symbols,

$$\forall a_1, a_2.\ a_1 \geq a_2 \Rightarrow F_b(\cdot|a_1 x_0)\ \text{FSD}\ F_b(\cdot|a_2 x_0), \tag{1}$$

where $F_b$ is the cumulative probability distribution for $b$, and $x_0$ is any assignment of values to the other predecessors of $b$.[1] An equivalent statement is that $\Pr(b \geq b_0 | a x_0)$ is non-decreasing in $a$, for any values of $b_0$ and $x_0$.

A negative influence is defined analogously with the appropriate changes in sign. When a non-monotonic or unknown probabilistic dependence holds, we assign a "?" influence. Independence is denoted by a zero influence, by convention represented implicitly in the absence of a link. For further motivation and implications of this definition, see [Wellman, 1990a].

### 3.2 Deterministic Relations

Qualitative influences on deterministic variables are simply functional monotonicity constraints. Let $b$ be a deterministic variable with predecessors $a$ and $x$, that is, $b = f(a, x)$ for some function $f$. Then $a$ positively influences $b$ iff, for any $a_1 > a_2$ and $x_0$, $f(a_1, x_0) > f(a_2, x_0)$. When $f$ is differentiable with respect to $a$, we can express this as an inequality on the partial derivative, $\frac{\partial b}{\partial a} > 0$.

If we regard deterministic functions as an extreme class of conditional probability distributions, we see that the definition of influences on deterministic variables is a special case of the probabilistic definition (1). Thus, any sound inference procedure for QPNs will produce sound conclusions in the presence of deterministic variables, ignoring the functional nature of the relations. The example of Section 2 demonstrated, however, that ignoring this information can lead to weakened conclusions, including spurious ambiguity.

Virtually all research in qualitative reasoning has been directed toward deterministic variables [Weld and de Kleer, 1989]. Even in a probabilistic setting, deterministic qualitative relations are likely to play a significant role in definitions, accounting relations, and constraints (for instance, in constrained optimization problems [Michelena and Agogino, 1989]). Conversely, the ability to express probabilistic relationships in otherwise deterministic models adds veridicality, since real-world problems invariably present elements of uncertainty.

### 3.3 A Note on Strictness and Continuity

The reader may find it curious that the deterministic relations are defined to be strictly monotonic, while the probabilistic definition employs non-strict inequalities. This practice follows the conventions of previous work: the original definitions of qualitative influences in QPNs [Wellman, 1990a] and the monotonicities used in dMIDs [Michelena and Agogino, 1989]. These conventions are not entirely arbitrary; the non-strict interpretation is notationally simpler and more broadly applicable in the probabilistic case, and strictness is required for invertibility in the deterministic.

---

[1] Without loss of generality, we can describe all the "other predecessors" in one possibly vector-valued variable: $x$.

Invertibility is necessary for the arc reversal operation in deterministic models, and can be qualitatively guaranteed only for strict monotone relations. In hybrid models, it makes sense to relax this requirement, allowing deterministic nodes to become probabilistic when their relations are inverted if the prerequisites do not hold. This suggests that it would be generally useful to admit both strict and non-strict qualitative relations, carefully distinguishing them and maintaining this information through network transformations. For notational simplicity, I adopt the convention in this paper that probabilistic relations are non-strict, while deterministic ones are strict, unless stated otherwise. Where strictness (or invertibility in general) is critical for the validity of an inference rule, the details are spelled out explicitly.

The use of partial derivatives to describe deterministic relationships presumes continuity and differentiability of the corresponding functions. Although the results presented here generally do not depend on these properties, I make use of them in proofs and illustrations for expository simplicity. The inference rules below could be justified by arguments based on differences as well as differentials. The only requirement is that the domain of every variable be ordinally scaled, so that monotonicity is a well-defined property.

## 4 Inference Rules

The probabilistic relations among any subset of a network's variables can be rendered direct via transformations composed of sequences of two basic operations [Shachter, 1988; Wellman, 1990b]. Node *reduction* is the process of removing a node from the network by averaging out its effects. The arc *reversal* operation changes the orientation of the directed edge, updating the probabilistic relation using Bayes's rule. A third operation, *deterministic node propagation* (DNP), removes links emanating from deterministic nodes. Each of these operations is associated with an update formula or inference rule describing how the qualitative relations need to be modified to reflect the changes in network structure.

Since qualitative deterministic relations are a special case of probabilistic ones, the inference rules for QPNs [Wellman, 1990a] are valid for hybrid networks. When the nodes of interest are *all* deterministic, the stronger dMID rules [Michelena and Agogino, 1989] apply. The rules below extend these sets to handle combinations of deterministic and probabilistic relations as well.

In the descriptions below, we consider a simple network fragment with nodes $a$, $c$, and $d$ ($b$ is saved for the discussion of synergy in Section 5). Node $c$ has a link to node $d$, and $a$ optionally has a link to each of the others. The situation is depicted in Figure 4.

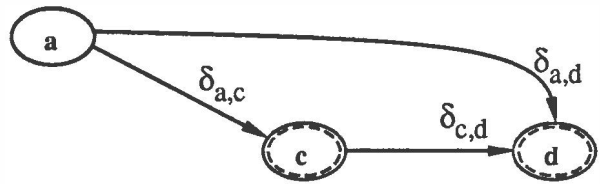

Figure 4: The simple network fragment used to illustrate the inference rules.

### 4.1 Deterministic Node Propagation

When a node is functionally determined by its predecessors, its outgoing arcs may be deleted via the operation of deterministic node propagation [Shachter, 1988]. The idea is that since the predecessors completely describe the node, it would be valid to describe the node's relation to its successors directly in terms of these predecessors. Therefore, the edge removal in DNP is accompanied by an update of the relation from the node's predecessors to its successors, adding new links if necessary.

Let $c$ be a deterministic node. Therefore, $c = f(a, x)$, where $a$ is the predecessor of interest and $x$ denotes the "other" predecessors. We consider first the case where its successor $d$ is also deterministic. Since $a$ may also be a predecessor of $d$ (as in Figure 4), $d = g(a, c, y)$ in general, where $y$ denotes the other predecessors of $d$, which may overlap with $x$. Substituting the expression for $c$, we see that

$$d = g(a, c, y) = g(a, f(a, x), y) \equiv \hat{g}(a, x, y).$$

Thus, $d$ is functionally determined by the union of $c$'s predecessors and its own, excepting $c$.

The inference problem we face is how to determine the qualitative relation of $a$ on $d$ with $c$ factored out. The solution is provided by the expression for $d$'s partial derivative with respect to $a$. To distinguish the perspectives of $g$ and $\hat{g}$, I use the notation $\frac{\partial d}{\partial a}(z, z', \ldots)$ to indicate the partial derivative expressed in terms of the variables in $\{z, z', \ldots\}$. According to the chain rule,[2]

$$\frac{\partial d}{\partial a}(a) = \frac{\partial d}{\partial a}(a, c) + \frac{\partial c}{\partial a}(a) \frac{\partial d}{\partial c}(a, c). \quad (2)$$

The signs of the terms on the right-hand side of (2) are given by the qualitative influences holding in the network before DNP. The sign of the influence

---

[2]I omit the $x$ and $y$ arguments from equation (2) to avoid unnecessary clutter.

from $a$ to $d$ after the operation, therefore, can be expressed by

$$\delta'_{a,d} = \delta_{a,d} \oplus (\delta_{a,c} \otimes \delta_{c,d}). \quad (3)$$

When $d$ is probabilistic, it will generally remain so after DNP. Nevertheless, the same update equation applies in this case. This fact is a consequence of the analogous result for node conditionalization in regular QPNs [Wellman, 1990a]. Although the conditionalization operation is valid in QPNs only for nodes with at most one successor, this restriction can be waived in the case of deterministic nodes due to the stronger D-separation condition for independence. Since $c$ is a function of its predecessors, the dependence of $d$ on any other variable in the context of these predecessors must be the same as it was in the context of $c$ and its own predecessors.

Figure 5 displays the general result of deterministic node propagation in QPNs. Node $d$ is deterministic iff it was so before the operation.

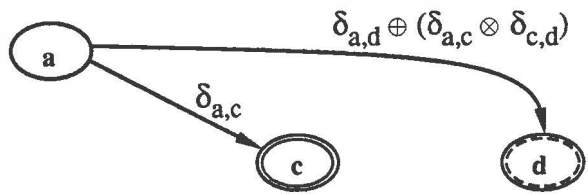

Figure 5: The network after deterministic node propagation. The doubled ellipse indicates that node $c$ is deterministic.

### 4.2 Reversal

An arc reversal operation transforms the network by flipping the orientation of a particular influence link and updating the incoming links of the incident variables. A link from $c$ to $d$ is eligible for reversal as long as there are no other paths from $c$ to $d$, in which case the operation would create a directed cycle.

Figure 6 depicts the structure of the network after reversal. The updated signs on the links ($\delta'$) are computed from the pre-reversal signs ($\delta$) according to Table 1. (Where the table indicates $\delta' = 0$, that link would be omitted from the network.) There are five cases, distinguished by the following factors:

- whether $c$ is deterministic (det) or probabilistic (prob)
- whether $d$ is deterministic or probabilistic, and
- whether $c$ has any predecessors.

The determinism of nodes $c$ and $d$ after reversal also depends on these factors.

The rules for deterministic $d$ also require that the function be invertible. If the qualitative influence of $c$ on $d$ is not strict and monotone, then the appropriate update procedure is to select the rule from Table 1 as if $d$ were probabilistic.

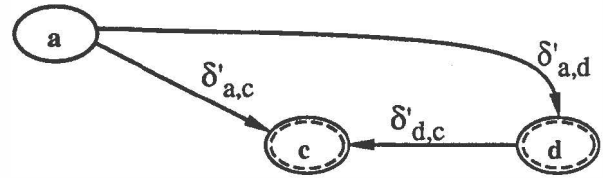

Figure 6: The network of Figure 4 after arc reversal. The new qualitative relations are given by Table 1.

Cases I and II are simply the reversal rules for dMIDs [Michelena and Agogino, 1989]. As they impose the strongest prerequisites on determinism of the variables, they yield the strongest results. At the other extreme, case V corresponds to the reversal rule for QPNs [Wellman, 1990a], which is valid for any reversible link. The results for any of the cases are at least as strong as the QPN results. Cases III and IV cover hybrid situations, where one of the nodes is deterministic and the other probabilistic.

Case III is actually not a reversal at all, but a complete removal of the link ($\delta'_{d,c} = 0$). It never makes sense to actually reverse the link from a deterministic node $c$ to a probabilistic node $d$, because, unlike the dMID case (I), node $d$ and its predecessors cannot be substituted for the original predecessors of $c$. Since these predecessors must remain anyway, adding a link from $d$ to already-deterministic $c$ would be superfluous. At best, the reversal would achieve the same results as when both nodes are probabilistic (case V). As deterministic node propagation (Section 4.1) dominates these results, for case III this operation should always be chosen over reversal.

An examination of the derivation of the rule for case I [Michelena and Agogino, 1989] reveals that the result does not depend on the determinism of $c$. Therefore, the updated signs for case IV are identical to those for the dMID rule. After reversal, $c$ becomes a deterministic function of $d$ and its predecessors, while $d$ turns into a probabilistic variable dependent on the union of $c$'s original predecessors with its own. The determinism of $c$ follows from invertibility of the original relation on $d$. The probabilistic nature of $d$ is a consequence of the stochastic relation between $c$ and its original predecessors. Without $c$, the remaining variables are insufficient to determine $d$ with certainty. And, in a departure from dMIDs (case II), an absence of original predecessors of $c$ does not permit us to separate $d$ from $a$, because the two variables may not be marginally independent. Thus, there is no special provision for $\text{pred}(c) = \emptyset$ in the case of probabilistic $c$.

| | $c$ | $d$ | pred($c$) | $\delta'_{d,c}$ | $\delta'_{a,c}$ | $\delta'_{a,d}$ | comment |
|---|---|---|---|---|---|---|---|
| I. | det | det | $\neq \emptyset$ | $\delta_{c,d}$ | $\ominus(\delta_{c,d} \otimes \delta_{a,d})$ | $\delta_{a,d} \oplus (\delta_{a,c} \otimes \delta_{c,d})$ | dMID |
| II. | det | det | $\emptyset$ | $\delta_{c,d}$ | $\ominus(\delta_{c,d} \otimes \delta_{a,d})$ | 0 | dMID |
| III. | det | prob | — | 0 | $\delta_{a,c}$ | $\delta_{a,d} \oplus (\delta_{a,c} \otimes \delta_{c,d})$ | DNP |
| IV. | prob | det | — | $\delta_{c,d}$ | $\ominus(\delta_{c,d} \otimes \delta_{a,d})$ | $\delta_{a,d} \oplus (\delta_{a,c} \otimes \delta_{c,d})$ | det↔prob |
| V. | prob | prob | — | $\delta_{c,d}$ | $\delta_{a,c} \oplus (\delta_{a,d} \otimes ?)$ | $\delta_{a,d} \oplus (\delta_{a,c} \otimes \delta_{c,d})$ | QPN |

Table 1: Rules for reversing the link from $c$ to $d$. Except for case IV, the deterministic or probabilistic nature of nodes is unchanged by the operation.

### 4.3 Reduction

Node reduction (also called removal or conditionalization) is the process of splicing a node out of the network, connecting its predecessors directly to its successors. In the simplest case, nodes without successors (*barren* nodes) can be summarily cut from the network. Nodes with successors can be reduced by reversing or deleting (via DNP) their outgoing links until they are barren, then removing them.

If node $c$ has a single successor, $d$, then the influence from any predecessor $a$ to $d$ after reducing c is
$$\delta'_{a,d} = \delta_{a,d} \oplus (\delta_{a,c} \otimes \delta_{c,d}).$$
Recall this expression is the same as the update for deterministic node propagation (3). In fact, when c is deterministic, this update is valid regardless of the number of successors. To see this, note that the link from $a$ to $c$ remains unchanged by DNP, and therefore a series of these operations for various successors can be performed independently. This invariance does not hold for arc reversal; the result of the first reversal generally affects subsequent ones. Thus probabilistic nodes cannot be directly reduced if they have more than a single successor.

After reducing $c$, node $d$ is deterministic iff both $c$ and $d$ were deterministic before the operation. Reducing a probabilistic node renders its successors probabilistic regardless of their former status.

### 4.4 Inference Rules: Discussion

Adopting special provisions for deterministic nodes strengthens the QPN inference rules in three primary ways.

1. An additional operation, deterministic node propagation, is available for eliminating links with less information loss than arc reversal.

2. Sharper results are obtained for arc reversals involving deterministic nodes. In Table 1, cases I-IV dominate case V in the sense that the conclusions are at least as strong for any assignment to the $\delta$s.

3. It is possible to directly reduce deterministic nodes with multiple successors.

The reduction of ambiguity demonstrated by the example of Section 2 is attributable to the last item.

The network of Figure 3b is the direct result of reducing $w$, chaining the influence from $z$ to both of $w$'s successors. If one of $w$'s outgoing links had to be reversed first (treating $w$ as probabilistic), the ambiguity of Figure 3a would be inevitable.

Given the augmented update rules, the inference task is to choose the appropriate sequence of operations to transform the network to answer specified queries. This choice is critical, as the strength of conclusions may vary depending on the transformation applied. I have addressed this issue in the context of QPNs [Wellman, 1990b], though the presence of deterministic variables presents some new questions. For example, when should reversal be chosen over deterministic node propagation? DNP is always preferred when $d$ is probabilistic (case III of Table 1), but neither operation is dominant in the other cases. Further work is required to resolve this and other inferential issues for QPNs with functional dependencies.

## 5 Qualitative Synergy

In the discussion thus far I have considered only qualitative influences. QPNs also include qualitative synergies describing the interaction of two variables in their influence on a third. Consider the network of Figure 7. A potential synergy $\delta_{\{x,y\},z} \in \{+,-,0,?\}$ exists between every pair of variables $x$ and $y$ with a common successor $z$.[3]

### 5.1 Definition

The definition for qualitative synergy is based on the concept of *supermodularity* [Topkis, 1978]. A bivariate function $z = f(x,y)$ is supermodular if it is more than additive in its arguments. Formally, for any $x_1 > x_2$, $y_1 > y_2$,
$$f(x_1, y_1) + f(x_2, y_2) \geq f(x_1, y_2) + f(x_2, y_1). \quad (4)$$
Analogous conditions with the inequality reversed or restricted to equality define submodularity and

---
[3]The synergies are not shown in the figure. Diagrams with explicit synergies (see [Wellman, 1990a]) get cluttered quadratically.

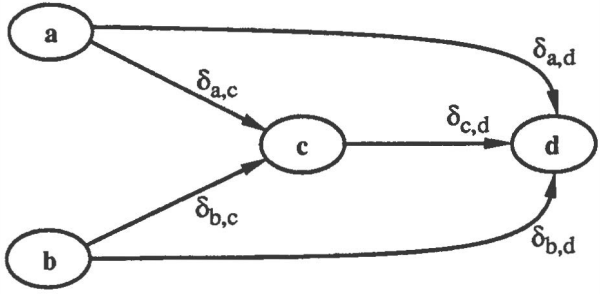

Figure 7: Qualitative synergies describe the interaction between influences on a common variable. In this example the relevant synergies are $\{a,b\}$ on $c$ and $\{a,b\}$, $\{a,c\}$, and $\{b,c\}$ on $d$.

## 5.2 Inference Rules

The question, then, is what inference rules are sanctioned for probabilistic and deterministic variables and combinations. Let us focus on the operation of reducing $c$ in the network fragment of Figure 7. If both $c$ and $d$ are probabilistic, the QPN update rule dictates the new synergy of $a$ and $b$ on $d$:

$$\begin{aligned}\delta'_{\{a,b\},d} &= \delta_{\{a,b\},d} \oplus (\delta_{\{a,b\},c} \otimes \delta_{c,d}) \\ &\oplus (\delta_{b,c} \otimes \delta_{\{a,c\},d}) \\ &\oplus (\delta_{a,c} \otimes \delta_{\{b,c\},d}). \end{aligned} \quad (6)$$

This rule is also valid if $d$ is the value node and synergies on $d$ are defined as supermodularity [Wellman, 1990a]. This follows from the invariance of utility under positive linear transforms.

However, the situation is different when $c$ or $d$ are deterministic. To see this, let us examine the case where both variables represent differentiable functions of their predecessors. Then the reduction operation transforms $d$ from a function of $a$, $b$, and $c$ to a function of $a$ and $b$ alone,

$$d = g(a,b,c,y) = g(a,b,f(a,b,x),y) \equiv \hat{g}(a,b,x,y).$$

The synergy of $a$ and $b$ on $d$ is expressed by the mixed second partial derivative, $\frac{\partial^2 d}{\partial a \partial b}$. We can relate the $\hat{g}$ version of this expression to the original $g$ representation using the chain rule.

$$\begin{aligned}\frac{\partial^2 d}{\partial a \partial b}(a,b) &= \frac{\partial^2 d}{\partial a \partial b}(a,b,c) + \frac{\partial^2 c}{\partial a \partial b}(a,b)\frac{\partial d}{\partial c}(a,b,c) \\ &+ \frac{\partial c}{\partial b}(a,b)\frac{\partial^2 d}{\partial a \partial c}(a,b,c) \\ &+ \frac{\partial c}{\partial a}(a,b)\frac{\partial^2 d}{\partial b \partial c}(a,b,c) \\ &+ \frac{\partial c}{\partial a}(a,b)\frac{\partial c}{\partial b}(a,b)\frac{\partial^2 d}{\partial c^2}(a,b,c). \quad (7)\end{aligned}$$

The first four additive terms on the right-hand side of (7) correspond exactly to the sign expressions of the four terms of the QPN synergy update equation (6). However, there is an additional term (arguments omitted),

$$\frac{\partial c}{\partial a}\frac{\partial c}{\partial b}\frac{\partial^2 d}{\partial c^2},$$

with no counterpart in (6). The first two factors of this term are described by the qualitative influences on $c$, but there is no qualitative relation corresponding to univariate second partial derivatives. Such a concept was not defined for QPNs, as it makes little sense for ordinally scaled variables. But it is not surprising that this is a factor in deterministic synergy, since the objects of that relation are in fact cardinally scaled variables. Indeed, Topkis [1978] has also shown that extension of submodularity depends on the convexity or concavity of the transformation function.

modularity, respectively. If $f$ is continuous and differentiable, (4) holds exactly when

$$\frac{\partial^2 z}{\partial x \partial y} \geq 0.$$

For deterministic variables, the synergy condition is simply supermodularity with respect to the specified predecessor variables. This is the definition provided for synergy on utility, a distinguished deterministic variable in QPNs [Wellman, 1990a].

Synergy on probabilistic variables is defined in terms of an inequality on differences in cumulative probability distributions for various combinations of conditioning variables [Wellman, 1990a]. The essential property of probabilistic synergy for our purposes is that it is equivalent to supermodularity of an expectation function. Specifically, $a$ and $b$ are synergistic on $c$ iff

$$E[\phi(c)|abx] \text{ is supermodular in } a,b \quad (5)$$

for any monotone transform $\phi$ and any assignment to the other predecessors $x$.

The requirement that (5) hold for all monotone transforms is a strong one, but it is precisely this condition that enables us to define synergy for merely ordinally scaled variables. The deterministic synergy condition, in contrast, is not invariant under monotone transforms, and therefore makes sense only for cardinally scaled variables. This departs also from qualitative influences, which are robust to monotone transforms in both the deterministic and probabilistic cases.

Thus, unlike the situation with influences, deterministic synergy is not a special case of its probabilistic counterpart. It is actually weaker in the sense that it permits us to presume a cardinal scale. However, it is stronger in another respect, namely that it mandates a functional dependency. Because neither relation subsumes the other, special treatment of deterministic variables for reasoning about synergy is required for soundness if the special interpretation of deterministic synergy is adopted.

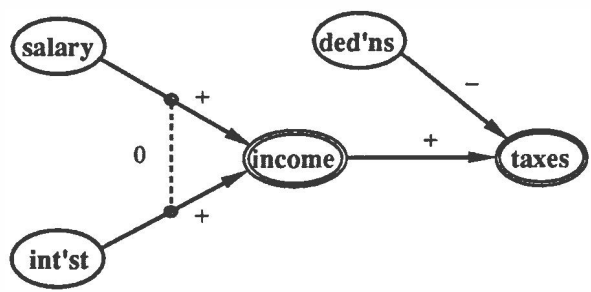

Figure 8: A QPN for the simplified tax-planning problem. The dashed line indicates a synergy of sign zero.

### 5.3 Example

Consider the simple tax-planning model of Figure 8. We define *income* as the sum of *salary* and *interest*, and assert that *taxes* are a function of *income* and *deductions*. *Salary* and *interest* have a zero synergy on *income*, as the combination function is additive, or modular.

Suppose we wish to determine the synergistic relation of *salary* and *interest* on *taxes*. This relation can be rendered direct by reducing *income* from the network. Because all synergies in the original network are zero, the first four terms of (7) drop out of the equation. The qualitative synergy of concern, therefore, depends entirely on whether the relation of *income* on *taxes* is concave or convex, that is, whether taxes are regressive or progressive. Unfortunately, this information cannot be expressed by qualitative influences and synergies alone.

It seems reasonable, then, that further investigation of qualitative deterministic synergy should be preceded by incorporation of univariate second-order qualitative relations. These relations represent natural concepts (concavity or convexity), and can also be propagated through graphical transformations (Nestor Michelena, personal communication). Establishing a probabilistic analog of these and developing methods to take advantage of their decision-theoretic implications are subjects for future work.

## 6 Conclusion

The foregoing analysis has demonstrated that augmenting QPNs to identify and exploit functional dependencies can strengthen inference in hybrid networks of deterministic and probabilistic variables. Moreover, much of the improvement can be realized by simple modifications to existing graphical inference rules. I have described these modifications in detail for qualitative influences and pointed out that carrying out a similar exercise for qualitative synergy will require the construction of new second-order qualitative relations.

Exploiting functional dependencies is likely to prove profitable for other representations based on probabilistic constraints. Any such scheme presents the potential for information loss when the constraints expressible in the specified language are not closed under the transformation operations [Fertig and Breese, 1989; Wellman, 1990b]. Functional dependencies, because they restrict the allowable interactions among variables, can significantly reduce this information loss in some cases.